\documentclass[11pt]{article}
\usepackage{scrextend}
\usepackage{eacl2017}
\usepackage{times}
\usepackage{url}
\usepackage{latexsym}
\usepackage{graphicx}
\usepackage{caption}
\usepackage{enumitem}
\usepackage{subfig}
\usepackage{amssymb}
\usepackage{amsmath}
\usepackage{svg}
\usepackage{multirow}
\usepackage{float}
\usepackage{tablefootnote}
\setsvg{inkscape = inkscape -z -D}
\setsvg{svgpath = images/}
\eaclfinalcopy

\newcommand\blfootnote[1]{%
  \begingroup
  \renewcommand\thefootnote{}\footnote{#1}%
  \addtocounter{footnote}{-1}%
  \endgroup
}

\title{Fine-Grained Entity Type Classification by Jointly Learning Representations and Label Embeddings}

\author{Abhishek, Ashish Anand \and Amit Awekar\\
  Department of Computer Science and Engineering\\
  Indian Institute of Technology Guwahati \\
  Assam, India - 781039 \\
  {\tt \{abhishek.abhishek, anand.ashish, awekar\}@iitg.ernet.in} \\
}
\date{}

\begin{document}
\maketitle
\begin{abstract}
  Fine-grained entity type classification (FETC) is the task of classifying an entity mention to a broad set of types. Distant supervision paradigm is extensively used to generate training data for this task. However, generated training data assigns same set of labels to every mention of an entity without considering its local context. Existing FETC systems have two major drawbacks: assuming training data to be noise free and use of hand crafted features. Our work overcomes both drawbacks. We propose a neural network model that jointly learns entity mentions and their context representation to eliminate use of hand crafted features. Our model treats training data as noisy and uses non-parametric variant of hinge loss function. Experiments show that the proposed model outperforms previous state-of-the-art methods on two publicly available datasets, namely \textsc{Figer(gold)} and \textsc{bbn} with an average relative improvement of 2.69\% in micro-F1 score. Knowledge learnt by our model on one dataset can be transferred to other datasets while using same model or other FETC systems. These approaches of transferring knowledge further improve the performance of respective models.

\end{abstract}
\section{Introduction}

Entity type classification is the task for assigning types or labels such as \textit{organization}, \textit{location} to entity mentions in a document. This classification is useful for many natural language processing (NLP) tasks such as relation extraction~\cite{mintz2009distant}, machine translation~\cite{koehn2007moses}, question answering~\cite{lin2012no} and knowledge base construction~\cite{dong2014knowledge}. 

There has been considerable amount of work on Named Entity Recognition (NER)~\cite{collins1999unsupervised,tjong2003introduction,ratinov2009design,manning2014stanford}, which classifies entity mentions into a small set of mutually exclusive types, such as \textit{Person}, \textit{Location}, \textit{Organization} and \textit{Misc}. However, these types are not enough for some NLP applications such as relation extraction, knowledge base construction (KBC) and question answering. In relation extraction and KBC, knowing fine-grained types for entities can significantly increase the performance of the relation extractor~\cite{ling2012fine,koch2014type,AAAI1510049} since this helps in filtering out candidate relation types that do not follow the type constrain. Fine-grained entity types provide additional information while matching questions to its potential answers and significantly improves performance~\cite{dong2015hybrid}. For example, Li and Roth~\shortcite{li2002learning} rank questions based on their expected answer types (will the answer be \textit{food}, \textit{vehicle} or \textit{disease}).  

\begin{figure}[h]
  \centering
  \includesvg[width = 1.0\columnwidth]{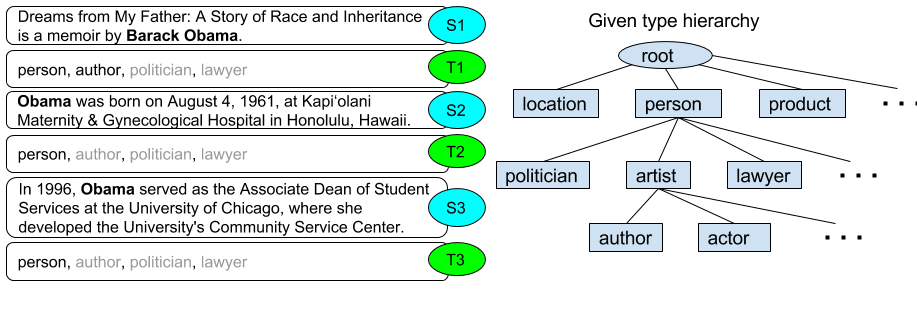}
  \caption{Noise introduced via distant supervision process. S1-S3 indicates sentences where only a subset of labels for entity mention (bold typeface) are relevant given context, highlighted in T1-T3.}
\label{fig:distant}
\end{figure}

Typically, FETC systems use over hundred labels,  arranged in a hierarchical structure. An important aspect of FETC is that based on local context, two different mentions of same entity can have different labels. We illustrate this through an example in Figure~\ref{fig:distant}. All three sentences \emph{S1}, \emph{S2}, and \emph{S3} mention same entity  \emph{Barack Obama}. However, looking at the context, we can infer that \emph{S1} mentions Obama as a \emph{person/author}, \emph{S2} mentions Obama only as a \emph{person}, and \emph{S3} mentions Obama as a \emph{person/politician}.

Available training data for FETC has noisy labels. Creating manually annotated training data for FETC is time consuming, expensive, and error prone. Note that, a human annotator will have to assign a subset of correct labels from a set of around hundred labels for each entity mention in the corpus. Existing FETC systems use distant supervision paradigm~\cite{craven1999constructing} to automatically generate training data. Distant supervision maps each entity in the corpus to knowledge bases such as Freebase~\cite{bollacker2008freebase}, DBpedia~\cite{auer2007dbpedia}, YAGO~\cite{suchanek2007yago}. This method assigns same set of labels to all mentions of an entity across the corpus. For example, \emph{Barack Obama} is a person, politician, lawyer, and author. If a knowledge base has these four matching labels for Barack Obama, then distant supervision assigns all of them to every mention of Barack Obama. Training data generated with distant supervision will fail to distinguish between mentions of Barack Obama in sentences \emph{S1}, \emph{S2}, and \emph{S3}.

Existing FETC systems have one or both of following drawbacks:
\begin{enumerate}[topsep=1pt,itemsep=2pt,partopsep=0ex,parsep=0ex]
  \item Assuming training data to be noise free~\cite{ling2012fine,spaniol2012hyena,yogatama2015embedding,shimaoka2016attentive}
  \item Use of hand crafted features~\cite{ling2012fine,spaniol2012hyena,yogatama2015embedding,AFET}
\end{enumerate}
We have observed that for real world datasets, more than twenty five percent of training data has noisy labels. First drawback propagates this noise in training data to the FETC model.  To extract hand crafted features various NLP tools are used. Since errors inevitably exist in such tools, the second drawback propagates errors of these tools to FETC model.

We propose a neural network based model to overcome the two drawbacks of existing FETC systems. First, we separate training data into \textit{clean} and \textit{noisy} partitions using the same method as in AFET system~\cite{AFET}. For these partitions, we use simple yet effective non-parametric variant of hinge loss function while training. To avoid use of hand crafted features, we learn representations for given entity mention and its context. 

Additionally, we investigate effectiveness of using transfer learning~\cite{pratt1993discriminability} for FETC task both at feature and model level. We show that feature level transfer learning can be used to improve performance of other FETC system such as AFET, by up to 4.5\% in micro-F1 score. Similarly, model level transfer learning can be used to improve performance of the same model using different dataset by up to 3.8\% in micro-F1 score.

Our contributions can be summarized as follows:
\begin{enumerate}[topsep=1pt,itemsep=2pt,partopsep=0ex,parsep=0ex]
  \item We propose a simple neural network model that learns representations for entity mention and its context, and incorporate noisy label information using a variant of non-parametric hinge loss function. Experimental results on two publicly available datasets demonstrate the effectiveness of proposed model, with an average relative improvement of 2.69\% in micro-F1 score.
  \item We investigate the use of feature level and model level transfer-learning strategies in the domain of the FETC task. The proposed transfer learning strategies further improve the state-of-the-art on BBN dataset by 3.8\% in micro-F1 score.
\end{enumerate}

\section{Related Work}
Ling et al.~\shortcite{ling2012fine} proposed the first system for FETC task, which used 112 overlapping labels. They used linear classifier perceptron for multi-label classification. Yosef et al.~\shortcite{spaniol2012hyena} used multiple binary SVM classifiers in a hierarchy, to classify an entity mention to a set of 505 types. While the initial work assumed that all labels present in a training dataset for an entity mention are correct, Gillick et al.~\shortcite{gillick2014context} introduced context dependent FETC and proposed a set of heuristics for pruning labels that might not be relevant given the entity mention's local context. Yogatama et al.~\shortcite{yogatama2015embedding} proposed an embedding based model where user-defined features and labels were embedded into a low dimensional feature space to facilitate information sharing among labels. 

Shimaoka et al.~\shortcite{shimaoka2016attentive} proposed an attentive neural network model that used LSTMs to encode entity mention's context and used an attention mechanism to allow the model to focus on relevant expressions in the entity mention's context. However, the model assumed that all labels obtained via distant supervision are correct. In contrast, our model does not assume that all labels are correct. To learn entity representation, we propose a scheme which is simpler yet more effective. 

\begin{figure}
  \centering
  \subfloat[$\alpha$ models label-label correlation. Higher the $\alpha$, lower is the margin between non-correlation labels.]{\includesvg[width=0.48\columnwidth]{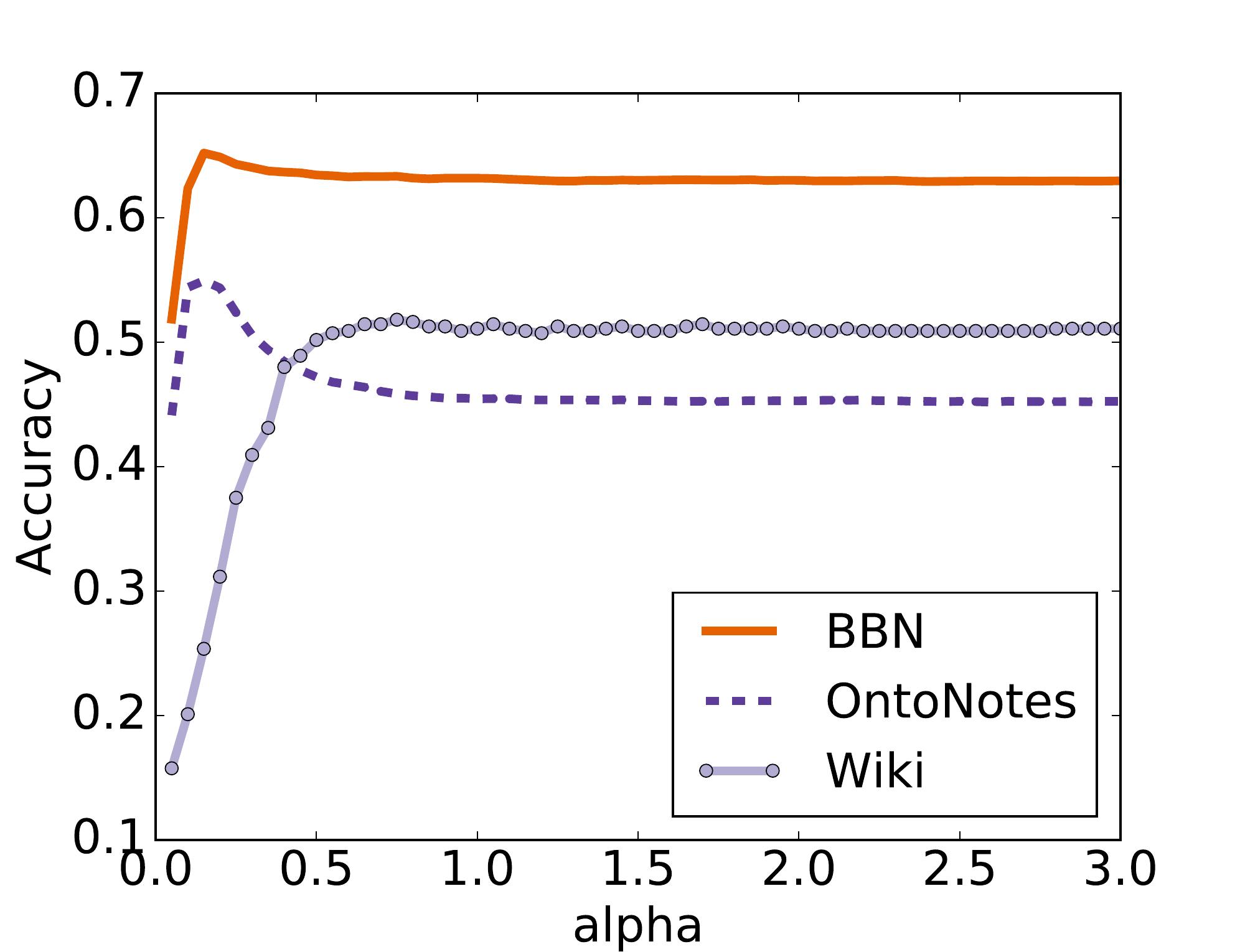}\label{fig:f1}}
  \hfill
\subfloat[During inference, labels above this threshold are predicted as positive.]{\includesvg[width=0.48\columnwidth]{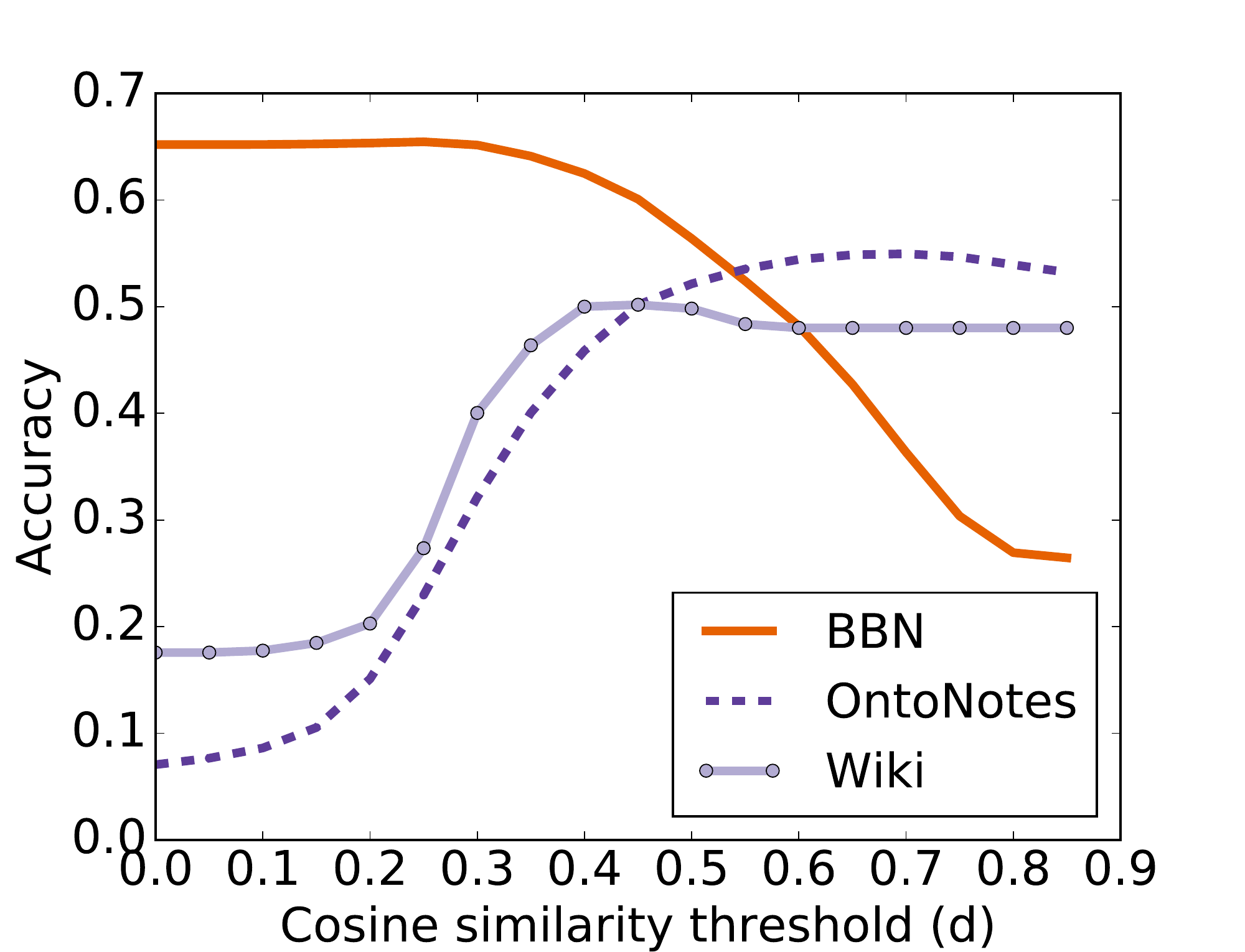}\label{fig:f2}}
  \caption{Effect of change of parameters on AFET's performance. }
\label{fig:AFET-parameters-variation}
\end{figure}

Most recently, Ren et al.~\shortcite{AFET} have proposed AFET, an FETC system. AFET separates the loss function for \textit{clean} and \textit{noisy} entity mentions. AFET uses label-label correlation information obtained by given data in its parametric loss function (model parameter $\alpha$). During inference, AFET uses a threshold to separate positive types from negative types (similarity threshold parameter $d$). However, AFET's loss function is sensitive to change in parameters, which are data dependent. Figure~\ref{fig:AFET-parameters-variation} shows the effect of parameter $\alpha$ and $d$, on AFET performance evaluated on different datasets. In contrast, our model uses a simple yet effective variant of hinge loss function. This function does not need to tune the similarity threshold. 

Transfer learning is well applied to many NLP applications, such as cross-domain document classification~\cite{shi2010cross}, multi-lingual word clustering~\cite{tackstrom2012cross} and sentiment classification~\cite{mou2015distilling}. Initialization of word vectors with pre-trained word vectors in neural network models can be considered as one of the best example of transfer learning in NLP\@. Wang et al.~\shortcite{wang2015transfer} provide a broad overview of transfer learning techniques used for language processing. 

\section{The Proposed Model}
\begin{figure}[!h]
  \centering
  \includesvg[width = 1.0\columnwidth]{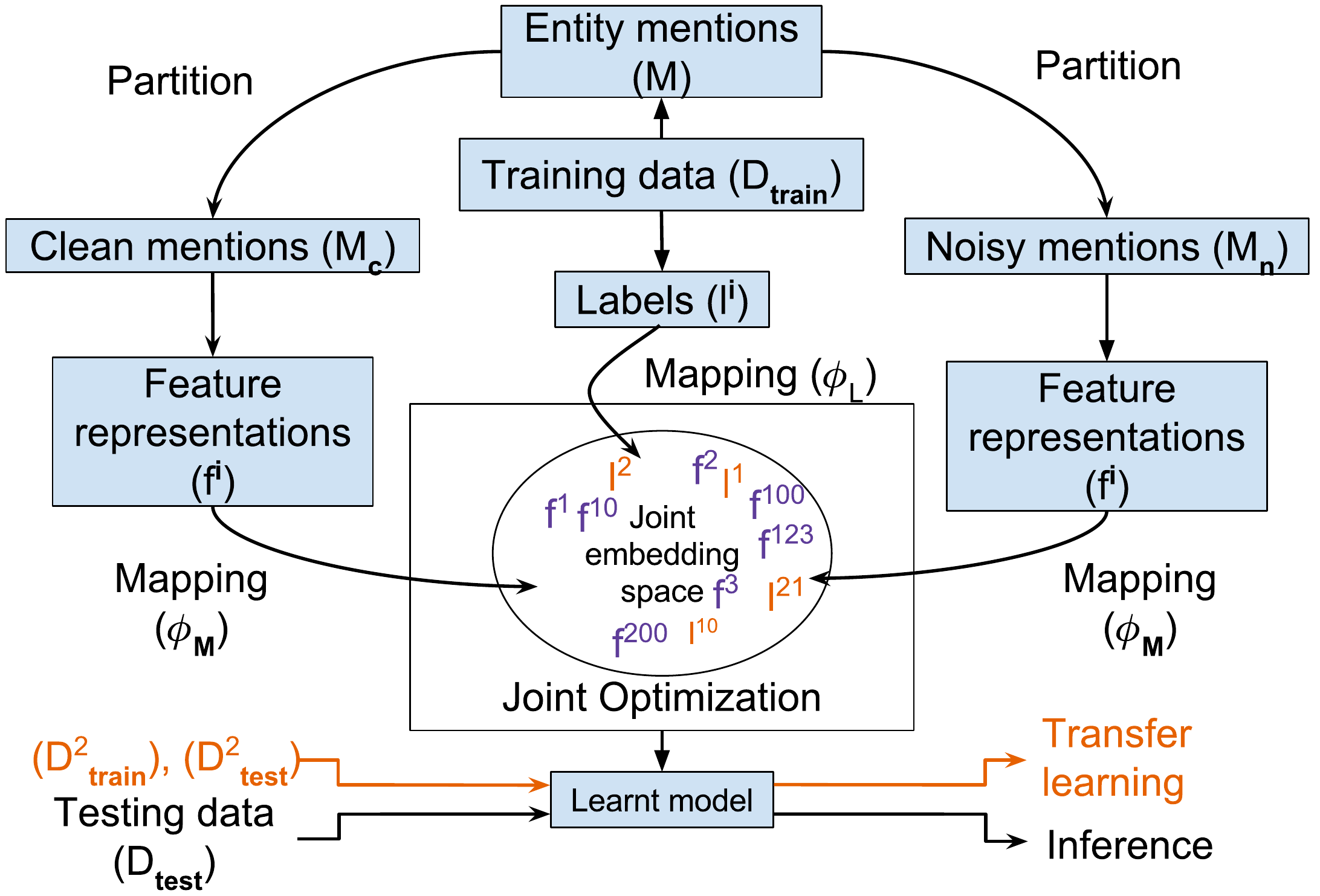}
  \caption{The system overview.\label{fig:system-diagram}}
\end{figure}
\begin{figure*}[t]
  \centering
  \includesvg[width = 1.0\textwidth]{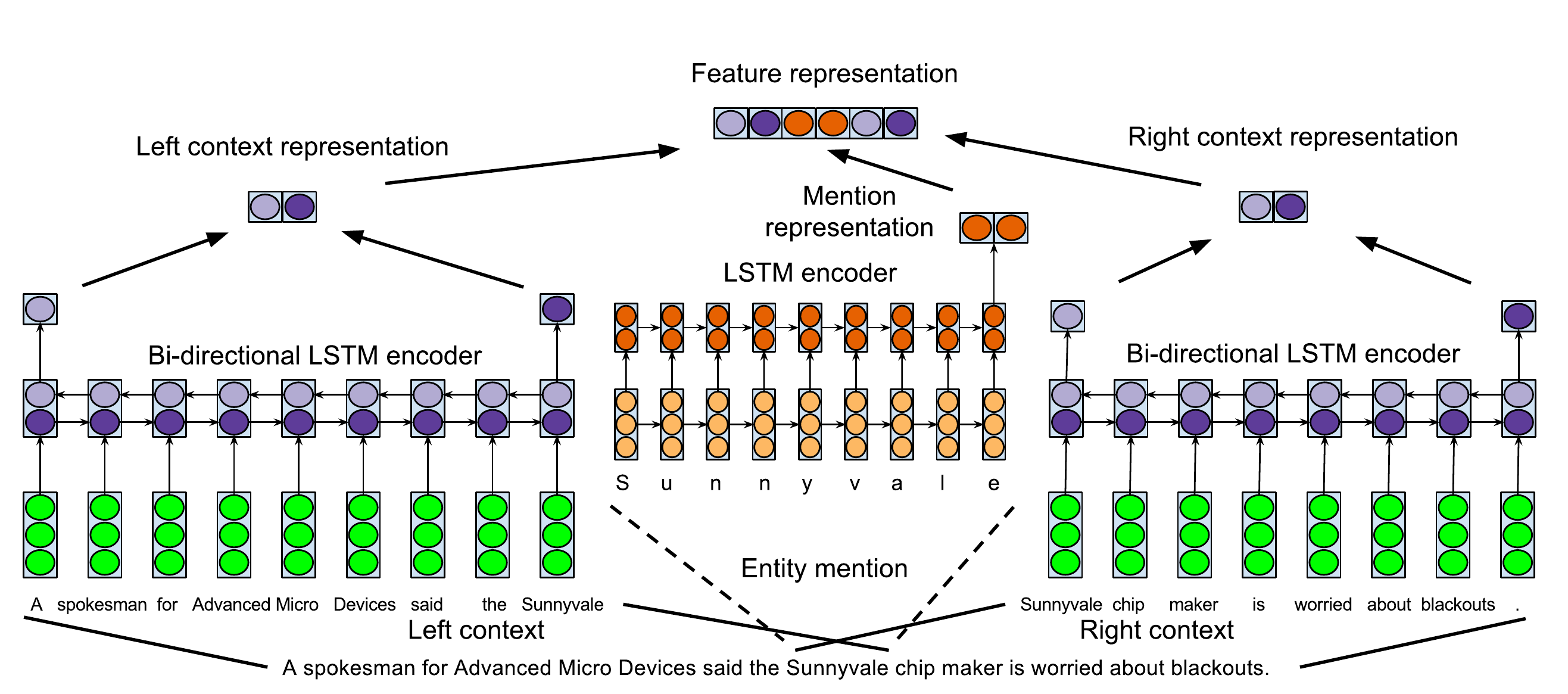}
  \caption{The architecture of the proposed model.\label{fig:model-diagram}}
\end{figure*}

\subsection{Problem description}
Our task is to automatically classify type information of entity mentions present in natural language sentences. Figure~\ref{fig:system-diagram} shows a general overview of our proposed approach.\\
\noindent
\textbf{Input:} The input to the model is a training and testing corpus consisting of a set of sentences on which entity mentions have been identified. In training corpus, every entity mention will have corresponding labels according to a given hierarchy. Formally, a training corpus $\mathcal{D}_{train}$ consists of a set of sentences, $\mathcal{S} = \{ s^{i}\}_{i=1}^{N}$. Each sentence $s^{i}$ will have one or more entity mentions denoted by $m_{j,k}^{i}$, where $j$ and $k$ denotes indices of start and end tokens, respectively. Set $\mathcal{M}$ consists of all the entity mentions $m_{j,k}^{i}$. For every entity mention $m_{j,k}^{i}$, there will be a corresponding label vector $l_{j,k}^{i} \in \{0,1\}^{K}$, which is a binary vector, where $l_{j,k_{t}}^{i} = 1$ if $t^{th}$ type is true otherwise it will be zero. $K$ denotes the total number of labels in a given hierarchy $\Psi$. Testing corpus $\mathcal{D}_{test}$ will only contain sentences and entity mentions.\\
\textbf{Output:} For entity mentions in testing corpus $\mathcal{D}_{test}$, predict their corresponding labels. 

\subsection{Training set partition}
Similar to AFET, we partition the mention set $\mathcal{M}$ of training corpus $\mathcal{D}_{train}$ into two parts, a set $\mathcal{M}_{c}$, consisting only of \textit{clean} entity mentions and a set $\mathcal{M}_{n}$, consisting only of \textit{noisy} entity mentions. An entity mention $m_{j,k}^{i}$ is said to be \textit{clean} if its labels $l_{j,k}^{i}$ belong to only a single path (not necessary to be leaf) in the hierarchy $\Psi$, that is its labels are not ambiguous; otherwise, it is \textit{noisy}. For example, as per hierarchy given in figure~\ref{fig:distant}, an entity mention with labels \emph{person}, \emph{artist} and \emph{politician} will be considered as \textit{noisy}, whereas entity mention with labels \emph{person}, \emph{artist} and \emph{actor} will be considered as \textit{clean}. 

\subsection{Feature representations}
\textbf{Mention representation:} This representation captures information about entity mention's morphology and orthography. We decompose an entity mention into character sequence, and use a vanilla LSTM encoder~\cite{hochreiter1997long} to encode character sequences to a fixed dimensional vector. Formally, for entity mention $m_{j,k}^{i}$, we decompose it into a sequence of character tokens $c_{j,k_{1}}^{i}$, $c_{j,k_{2}}^{i}$, $\dots$,$c_{j,k_{|m_{j,k}^{i}|}}^{i}$, where $|m_{j,k}^{i}|$ denotes the total number of characters present in the entity mention. For entity mention containing multiple tokens, we join these tokens with a space in between tokens. Every character will have corresponding vector representation in a lookup table for characters. The character sequence is then fed one by one to a LSTM encoder, and the final output is used as a feature representation for entity mention $m_{j,k}^{i}$. We denote this process by a function $F_{m}:\mathcal{M} \to \mathbb{R}^{D_{m}}$, where $D_{m}$ is the number of dimensions for mention representation. The whole process is illustrated in figure~\ref{fig:model-diagram} (Mention representation).\\
\textbf{Context representation:} This representation captures information about the context surrounding the entity mention. Context representation is further divided into two parts, left and right context representation. The left context consists of a sequence of tokens within a sentence from the start of a sentence till the last token of entity mention. The right context consists of a sequence of tokens from the start of entity mention till the end of a sentence. We use bi-directional LSTM encoders \cite{graves2013speech} to encode token level sequences of both context to a fixed dimensional vector. Formally, for an entity mention $m_{j,k}^{i}$ present in a sentence $s^{i}$, decompose $s^{i}$ into a sequence of tokens $s_{1}^{i}$, $s_{2}^{i}$, $\dots$, $s_{k}^{i}$ for the left context, and $s_{j}^{i}$, $s_{j+1}^{i}$, $\dots$, $s_{|s^{i}|}^{i}$ for the right context, where $|s^{i}|$ denotes the number of tokens in the sentence. Every token will have a corresponding vector representation in a lookup tables for token. The token sequence is then fed one by one to a bi-directional LSTM encoder, and the final output will be used as feature representation. We denote this whole process by function $F_{lc}:(\mathcal{M,S}) \to \mathbb{R}^{D_{lc}}$ for computing left context and $F_{rc}:(\mathcal{M,S}) \to \mathbb{R}^{D_{rc}}$ for computing right context. $D_{lc}$ and $D_{rc}$ are the number of dimensions for the left context and the right context representation, respectively. The whole process is illustrated in figure~\ref{fig:model-diagram} (Left and right context representation). 

The context representation described above is slightly different from what was proposed in~\cite{shimaoka2016attentive}, here we include entity mention tokens within both left and right context, to explicitly encode context relative to an entity mention.

In the end, we concatenate entity mention and its context representation into a single $D_{f}$ dimensional vector, where $D_{f} = D_{m} + D_{lc} + D_{rc}$. This complete process is denoted by a function $F:(\mathcal{M,S}) \to \mathbb{R}^{D_{f}}$ given by:
\begin{equation}
  \resizebox{1.0\columnwidth}{!}{$F(m_{j,k}^{i}, s^{i}) = F_{m}(m_{j,k}^{i}) \oplus F_{lc}(m_{j,k}^{i},s^{i}) \oplus F_{rc}(m_{j,k}^{i},s^{i})$}
\end{equation}
where $\oplus$ denotes vector concatenation. 
For brevity, we will now omit the use of subscript $j,k$ from $m_{j,k}^{i}$ and $l_{j,k}^{i}$, and will use $f^{i}$ to denote feature representation for entity mention and its context obtained via equation 1. 
\subsection{Feature and label embeddings}
Similar to Yogatama et al.~\shortcite{yogatama2015embedding} and Ren et al.~\shortcite{AFET}, we embed feature representations and labels in a same dimensional space such that an object is embedded closer to the objects that share similar types than the objects that do not. Formally, we are trying to learn linear mapping functions $\phi_{\mathcal{M}}: \mathbb{R}^{D_{f}} \to \mathbb{R}^{D_{e}}$ and $\phi_{\mathcal{L}}: \mathbb{R}^{D_{K}} \to \mathbb{R}^{D_{e}}$, where $D_{e}$ is the size of embedding space. These mappings are given by: 
\begin{equation}
  \phi_{\mathcal{M}}(f^{i}) = f^{i^{T}}U; \; \phi_{\mathcal{L}}(l^{i}_{t}) = l^{i^{T}}_{t}V
\end{equation}
where, $U \in \mathbb{R}^{D_{f}\times D_{e}}$ and $V \in \mathbb{R}^{D_{K}\times D_{e}}$ are projection matrices for features representations and type labels respectively and $l^{i}_{t}$ is one-hot vector representation for label $t$.\\ 
We assign a score to each label type $t$ and feature vector as a dot product of their embeddings. Formally, we denote a score as:
\begin{equation}
  s(f^{i},l^{i}_{t}) = \phi_{\mathcal{M}}(f^{i}) \cdot \phi_{\mathcal{L}}(l^{i}_{t}) 
\end{equation}
\subsection{Optimization}
We use two different loss functions to model \textit{clean} and \textit{noisy} entity mentions. For \textit{clean} entity mentions, we use a hinge loss function. The intuition is simple: maintain a margin, centered at zero, between positive and negative type scores. The scores are computed by similarity between an entity mention and label types (eq\@. 3). Hinge loss function has two advantages. First, it intuitively seprates positive and negative labels during inference. Second, it is independent of data dependent parameter. Formally, for a given entity mention $m^{i}$ and its label $l^{i}$ we compute the associated loss as given by:
\begin{align}
  L_{c}(m^{i}, l^{i}) &= \sum_{t \in \gamma} \max(0, 1 - s(m^{i},l^{i}_{t}))  \notag \\
                      &+ \sum_{t \in \bar{\gamma}} \max(0, 1 + s(m^{i}, l^{i}_{t}))
\end{align}
where $\gamma$ and $\bar{\gamma}$ are set of indices that have positive and negative labels respectively. 

For \textit{noisy} entity mentions, we propose a variant of a hinge loss where, like $L_{c}$, score for all negative labels should go below $-1$. However, for positive labels, as we don't know which labels are relevant to entity mention's local context, we propose that the maximum score from the set of given positive labels should be greater than one. This maintains a margin between all negative types and the most relevant positive type. Formally, \textit{noisy} label loss, $L_{n}$ is defined as:

\begin{align}
  L_{n}(m^{i}, l^{i}) &= \sum_{t \in \bar{\gamma}} \max(0, 1 + s(m^{i},l^{i}_{t})) \notag \\
  &+ \max(0, 1 - s(m^{i},l^{i}_{t^{*}}));  \notag \\
  t^{*} &= \arg\max_{t \in \gamma}s(m^{i},l^{i}_{t})
\end{align}
Again, using this loss function makes it intuitive to set a threshold of zero during inference. 

These loss functions are different from the loss functions used in~\cite{yogatama2015embedding,AFET} in a way that, we make strict absolute criteria to distinguish between positive and negative labels. Whereas in~\cite{yogatama2015embedding,AFET} positive labels should have a higher score than negative labels. As their scoring is relative, the final result varies on the threshold used to separate positive and negative labels.

To train the partitioned dataset together, we formulate the joint objective problem as:
\begin{align}
  \min_{\theta} O = \sum_{m \in \mathcal{M}_{c}}L_{c}(m,l) + \sum_{m \in \mathcal{M}_{n}}L_{n}(m,l)
\end{align}
where $\theta$ is the collection of all model parameters that needs to be learned. To jointly optimize the objective $O$, we use Adam~\cite{kingma2014adam}, a stochastic gradient-based optimization algorithm. 
\subsection{Inference}
For every entity mention in set $\mathcal{M}$ from $\mathcal{D}_{test}$, we perform a top-down search in the given type hierarchy $\Psi$, and estimate the correct type path $\Psi_{*}$. Starting from the tree root, we recursively compute the best type among node's children by computing its score with obtained feature representations. We select the node that has maximum score among other nodes. We continue this process till a leaf node is encountered or the score associated with a node falls below an absolute threshold zero. The thresold is fixed across all datasets used. 
\subsection{Transfer learning}
We want to investigate, whether the feature representations learnt for an entity mention are useful. We study what contribution these feature representations make to an existing feature engineering based method such as AFET\@.  We learn the proposed model on one training dataset, namely Wiki dataset, which has the highest number of entity mentions among other datasets and use this model to generate representations that is $F(m_{j,k}^{i}, s^{i})$ for another training and testing data. These representations, which are $D_{f}$ dimensional vectors, are used as feature for an existing state-of-the-art model, AFET, in place of the hand-crafted features that were originally used. AFET model is then trained using these feature representations. We call this as feature level transfer learning. On the other hand, we also evaluate model level transfer learning, where we initialize weights of LSTM encoders for a new dataset with the weights learnt from the model trained on another dataset, namely Wiki dataset.
\section{Experiments}
\subsection{Datasets used}
We evaluate the proposed model on three publicly available datasets, provided in a pre-processed tokenized format by Ren et al.~\shortcite{AFET}. Statistics of the datasets used in this work are shown in Table~\ref{tab:dataset-statistics}. The details of the datasets are as follows:\\
\textbf{Wiki/\textsc{Figer(gold)}:} The training data consists of Wikipedia sentences and was automatically generated in distant supervision paradigm, by mapping hyperlinks in Wikipedia articles to Freebase. The test data, mainly consisting of sentences from news reports, was manually annotated as described in~\cite{ling2012fine}.\\
\textbf{OntoNotes:} OntoNotes dataset consists of sentences from newswire documents present in OntoNotes text corpus~\cite{weischedel2013ontonotes}. DBpedia spotlight~\cite{daiber2013improving} was used to automatically link entity mention in sentences to Freebase. For this corpus, manually annotated test data was shared by Gillick et al.~\shortcite{gillick2014context}.\\
\textbf{BBN:} BBN dataset consists of sentences from Wall Street Journal articles and is completely manually annotated~\cite{weischedel2005bbn}.\\
Please refer to~\cite{AFET} for more details of the datasets. 
\begin{table}[]
  \centering
  \resizebox{\linewidth}{!}{%
  \begin{tabular}{llll}
  \hline
  Datasets                        & Wiki/\textsc{Figer(gold)}    & OntoNotes   & BBN     \\ \hline
  \# types                        & 128     & 89          & 47      \\ 
  \# training mentions            & 2690286 & 220398      & 86078   \\
  \# testing mentions             & 563     & 9603        & 13187   \\
  \% clean training mentions      & 64.58   & 72.61       & 75.92   \\
  \% clean testing mentions       & 88.28   & 94.00       & 100     \\
  \% pronominal testing mentions\tablefootnote{We considered an entity mention as pronominal, if all of its tokens have POS tag as pronoun.} & 0.00    & 6.78        & 0.00    \\
  Max hierarchy depth             & 2       & 3           & 2       \\ 
\end{tabular}%
}
\caption{Statistics of the datasets used in this work.}
\label{tab:dataset-statistics}
\end{table}

\subsection{Evaluation settings}


\begin{figure*}[t]
  \centering
  \includesvg[width = 1.0\textwidth]{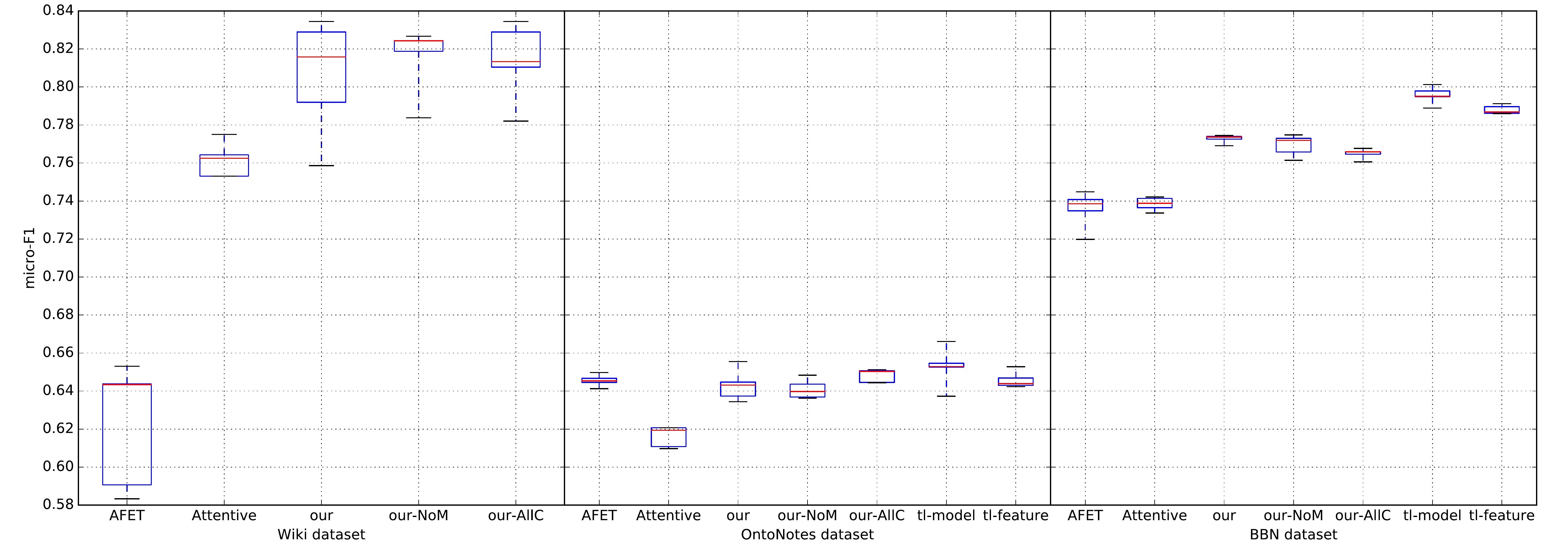}
  \caption{These box-plots show the performance of different baselines on validation set. The red line, boxes and whiskers indicate the median, quartiles and range.\label{fig:box_plots}}
\end{figure*}

\subsubsection{Baselines}
We compared the proposed model with state-of-the-art entity classification methods\footnote {Whenever possible, the baselines result are reported from~\cite{AFET}, otherwise we re-implemented baseline methods based on description available in corresponding papers.}: (1) \textbf{FIGER}~\cite{ling2012fine}; (2) \textbf{HYENA}~\cite{spaniol2012hyena}; (3) \textbf{AFET-NoCo}~\cite{AFET}: AFET without data based label-label correlation modeled in loss function; (4) \textbf{AFET-CoH}~\cite{AFET}: AFET with hierarchy based label-label correlation modeled in loss function; (5) \textbf{AFET}~\cite{AFET}; (6) \textbf{Attentive}~\cite{shimaoka2016attentive}: An attentive neural network based model. 

We compare these baselines with variants of our proposed model: (1) \textbf{our}: complete model; (2) \textbf{our-AllC} assuming all mentions are \textit{clean}; (3) \textbf{our-NoM} without mention representation. 
\subsubsection{Experimental setup}
We use Accuracy or Strict-F1 score, Macro-averaged F1 score, and Micro-averaged F1 score as metrics for evaluation. Existing methods for FETC use same measures \cite{ling2012fine,yogatama2015embedding,shimaoka2016attentive,AFET}. We removed entity mentions that do not have any label in training as well as test set. We also remove entity mentions that have spurious indices (i.e entity mention length of $0$).\footnote{The code to replicate the work is available at \url{https://github.com/abhipec/fnet}} For all the three datasets, we randomly sampled 10\% of the test set, and use it as a development set, on which we tune model parameters. The remaining 90\% is used for final evaluation. For all our experiments, we train each model using same hyperparameters five times and report their performance in terms of micro-F1 score on the development set as shown in Figure~\ref{fig:box_plots}. On Wiki dataset, we observed a large variance in performance as compared to other two datasets. This might be because of the fact that Wiki dataset has a very small development set. From each of these five runs, we pick the best performing model based on the development set and report its result on the test set. \\ 
\textbf{Hyperparameter setting:} All the neural network based models in this paper used 300 dimensional pre-trained word embeddings distributed by Pennington et al.~\shortcite{pennington2014glove}. The hidden-layer size of word level bi-directional LSTM was 100, and that of character level LSTM was 200. Vectors for character embeddings were randomly initialized and were of size 200. We use dropout with the probability of 0.5 on the output of LSTM encoders. The embedding dimension used was 500. We use Adam~\cite{kingma2014adam} as optimization method with learning rate of 0.0005-0.001 and mini-batch size in the range of 800 to 1500. The proposed model and some of the baselines were implemented using TensorFlow\footnote{\url{http://tensorflow.org/}} framework.

\begin{table*}[t]
  \centering
  \resizebox{\textwidth}{!}{%
  \begin{tabular}{|l|lll|lll|lll|}
  \hline
  \multirow{2}{*}{Typing methods}      & \multicolumn{3}{c|}{Wiki/\textsc{Figer(gold)}} & \multicolumn{3}{c|}{OntoNotes} & \multicolumn{3}{c|}{BBN}\\ \cline{2-10} 
                                                  & Acc.  & Ma-F1  & Mi-F1        & Acc.  & Ma-F1  & Mi-F1  & Acc.  & Ma-F1  & Mi-F1 \\ \hline
  \textbf{FIGER}\textsuperscript{*}~\cite{ling2012fine}              & 0.474 & 0.692 & 0.655         & 0.369 & 0.578 & 0.516 & 0.467 & 0.672 & 0.612 \\ 
  \textbf{HYENA}\textsuperscript{*}~\cite{spaniol2012hyena}          & 0.288 & 0.528 & 0.506         & 0.249 & 0.497 & 0.446 & 0.523 & 0.576 & 0.587 \\ 
  \textbf{AFET-NoCo}\textsuperscript{*}~\cite{AFET}                  & 0.526 & 0.693 & 0.654         & 0.486 & 0.652 & 0.594 & 0.655 & 0.711 & 0.716 \\ 
  \textbf{AFET-CoH}\textsuperscript{*}~\cite{AFET}                   & 0.433 & 0.583 & 0.551         & 0.521 & 0.680 & 0.609 & 0.657 & 0.703 & 0.712 \\ 
  \textbf{AFET}\textsuperscript{*}~\cite{AFET}                       & 0.533 & 0.693 & 0.664         & 0.551 & 0.711 & 0.647 & 0.670 & 0.727 & 0.735 \\ 
  \textbf{AFET}\textsuperscript{\textdagger \ddag}~\cite{AFET}                       & 0.509 & 0.689 &  0.653 & \textbf{0.553} & \textbf{0.712} & \textbf{0.646} &  0.683 & 0.744 & 0.747\\ 
  \textbf{Attentive}\textsuperscript{\textdagger}~\cite{shimaoka2016attentive} & 0.581 & 0.780 & 0.744 & 0.473 & 0.655 & 0.586 & 0.484 & 0.732 & 0.724 \\ \hline
  \textbf{our-AllC}\textsuperscript{\textdagger}                               & \textbf{0.662} & 0.805 & 0.770 & 0.514 & 0.672 & 0.626 & 0.655 & 0.736 & 0.752 \\ 
  \textbf{our-NoM}\textsuperscript{\textdagger}                                & 0.646 & 0.808 & 0.768 & 0.521 & 0.683 & 0.626 & 0.615 & 0.742 & 0.755 \\ 
  \textbf{our}\textsuperscript{\textdagger}                                    & 0.658 & \textbf{0.812} & \textbf{0.774} & 0.522 & 0.685 & 0.633 & 0.604 & 0.741 & 0.757 \\ \hline
  \textbf{model level transfer-learning}\textsuperscript{\textdagger}          & - & - & - & 0.531 & 0.684 & 0.637 & 0.645 & 0.784 & \textbf{0.795} \\ 
  \textbf{feature level transfer-learning}\textsuperscript{\textdagger}        & - & - & - & 0.471 & 0.689 & 0.635 & \textbf{0.733} & \textbf{0.791} & 0.792 \\ \hline 
\end{tabular}%
}
\caption{Performance analysis of entity classification methods on the three datasets.}
\label{tab:results}
\end{table*}

\subsection{Transfer learning}
In feature level transfer learning, we use the best performing proposed model trained on Wiki dataset to generate representations that is $D_{f}$ dimensional vector for every entity mention present in the train, development, and test set of the BBN and the OntoNotes dataset. Figure~\ref{fig:model-diagram} illustrates an example for the encoding process. Then we use these representations as a feature vector in place of the user-defined features and train the AFET model. Its hyper-parameters were tuned on the development set. These results are shown in table~\ref{tab:results} as \textbf{feature level transfer-learning}. 

In model level transfer learning, we use the learnt weights of LSTM encoders from the best performing proposed model trained on Wiki dataset and initialize the LSTM encoders of the same model with these weights while training on BBN and OntoNotes datasets. These results are shown in table~\ref{tab:results} as \textbf{model level transfer learning}. 
\subsection{Performance comparison and analysis}
Table~\ref{tab:results} shows the results of the proposed method, its variants and the baseline methods. \\
\textbf{Comparison with other feature learning methods:} The proposed model and its variants (\textbf{our-AllC, our-NoM}) perform better than the existing feature learning method by Shimaoka et al.~\shortcite{shimaoka2016attentive} (\textbf{Attentive}), consistently on all datasets. This indicates benefits of the proposed representation scheme and joint learning of representation and label embedding. \\
\textbf{Comparison with feature engineering methods:} The proposed model performs better than the existing feature engineered methods (\textbf{FIGER}, \textbf{HYENA}, \textbf{AFET-NoCo}, \textbf{AFET-CoH}) consistently across all datasets on Micro-F1 and Macro-F1 evaluation metrics. These methods do not model label-label correlation based on data.  In comparison with \textbf{AFET}, the proposed model outperforms AFET on Wiki and BBN dataset in terms of Micro-F1 evaluation metric. This indicates benefits of feature learning as well as data driven label-label correlation. We do a type-wise performance comparison on OntoNotes dataset in subsection 4.5.\\
\textbf{Comparison with variants of our model:} The proposed model performs better on all dataset as compared to \textbf{our-AllC} in terms of micro-F1 score. However, we find the performance difference on Wiki and OntoNotes dataset is not statistically significant. We investigated it further and found that across all three datasets, there exist only few entity types for which more than 85\% of entity mentions are noisy. These types consist of approximately 3-4\% of test set, and our model fails on these types (zero micro-F1 score). However, \textbf{our-AllC} performs relatively well on these types. Examples of such types are: \textit{/building}, \textit{/person/political\_figure}, \textit{/GPE/STATE\_PROVINCE}. This indicates two limitations of the proposed model. First, the separating of clean and noisy mentions based on the hierarchy has its own inherent limitation of assuming labels within a path are correct. Second, our model learns better if more clean examples are available at the cost of not learning very noisy types. We will try to address these limitations in our future work. Compared with \textbf{our-NoM}, the proposed model performs slightly better across all datasets in terms of micro-F1 score.\\
\textbf{Feature level transfer learning analysis:} We observed $4.5\%$ performance increase in micro-F1 score of AFET on BBN dataset, after replacing hand-crafted features with feature representations generated by the proposed model. This indicates usefulness of the learnt feature representations. However, if we repeat the same process with OntoNotes dataset, there is only a subtle change in performance. This is majorly because of the data distribution of OntoNotes dataset is different from that of Wiki dataset. This issue is discussed in the next subsection.\\
\textbf{Model level transfer learning analysis:} In model level transfer learning, sharing knowledge from similar dataset (Wiki to BBN) increases the performance by 3.8\% in terms of micro-F1 score. However, sharing knowledge from Wiki to OntoNotes dataset slightly increases the performance by 0.4\% in terms of micro-F1 score.
\blfootnote{\textsuperscript{*}These results are from \cite{AFET} that also uses 10\% of the test set as development set and the remaining for evaluation.}
\blfootnote{\textsuperscript{\ddag}We used the publicly available code distributed by Ren et al.~\shortcite{AFET}.}
\blfootnote{\textsuperscript{\textdagger}All of these results are on exact same train, development and test set.}
\subsection{Case analysis: OntoNotes dataset}
We observed three things; (i) all models perform relatively poor on OntoNotes dataset compared to their performance on other two datasets; (ii) the proposed model outperforms other models including AFET on the other two datasets, but gave worse performance on OntoNotes dataset; (iii) the two variants of transfer learning significantly improve performance of the proposed model on the BBN dataset but resulted in only a subtle performance change on OntoNotes dataset.

Statistics of the dataset (Table~\ref{tab:dataset-statistics}) indicates that presence of pronominal or other kinds of mentions are relatively higher in OntoNotes ($6.78\%$ in test set) than the other two datasets ($0\%$ in test set). Examples of such mentions are \textit{100 people}, \textit{It}, \textit{the director}, etc. Table~\ref{tab:random-entity-names} shows 20 randomly sampled entity mentions from test set of OntoNotes datasets. Some of these mentions are very generic and likely to be dependent on previous sentences. As all the methods use features solely based on the current sentence, they fail to transfer cross-sentence boundary knowledge. Removing pronominal mentions from test set increases the performance of all feature learning methods by around 3\%. 

\begin{table}[h]
  \centering
  \resizebox{\columnwidth}{!}{%
  \begin{tabular}{|ll|}
    \hline
   his            & thousands of angry people \\ 
   A reporter     & export competitiveness               \\ 
   Freddie Mac    & Messrs. Malson and Seelenfreund     \\ 
   the numbers    & Hollywood and New York               \\ 
   his explanation& April                        \\ 
   volatility     & This institution               \\ 
   their hands    & the 1987 crash               \\ 
   it             & January 4th                 \\ 
   Macau          & investment enterprises      \\ 
   France         & any means                   \\ \hline
    
\end{tabular}%
}
\caption{20 randomly sampled entity mentions present in the test set of OntoNotes dataset.}
\label{tab:random-entity-names}
\end{table}
Next we analyse where the proposed model is failing as compared to AFET\@. For this, we look at type-wise performance for the top-10 most frequent types in the OntoNotes test dataset. Results are shown in Table~\ref{tab:class-wise_OntoNotes}. Compared to AFET, the proposed model performs better in all types except \textit{other} in the top-10 frequent types. The \textit{other} type, which is dominant in test set ($42.6\%$ of entity mentions are of type \textit{other}) and is a collection of multiple broad subtypes such as \textit{product}, \textit{event}, \textit{art}, \textit{living\_thing}, \textit{food}. Performance of AFET significantly drops (\textit{AFET-NoCo}) when data-driven label-label correlation is ignored, which indicates that modeling data-driven correlation helps. However, as shown in Figure~\ref{fig:f1}, the use of label-label correlation depends on appropriate values of parameters which vary from one dataset to another.

\begin{table}[h]
  \centering
  \resizebox{\columnwidth}{!}{%
  \begin{tabular}{|l|c|lll|lll|}
  \hline
  \multirow{2}{*}{Label type}       &Support& \multicolumn{3}{c|}{our} & \multicolumn{3}{c|}{AFET} \\ \cline{3-8} 
                                    &       & Prec. & Rec.  & F-1           & Prec. & Rec.  & F-1   \\ \hline
  \textit{/other}                   & 42.6\%& 0.838 & 0.809 & 0.823         & 0.774 & 0.962 & \textbf{0.858}\\ \hline
  \textit{/organization}            & 11.0\%& 0.588 & 0.490 & \textbf{0.534}& 0.903 & 0.273 & 0.419         \\ \hline
  \textit{/person}                  & 9.9\% & 0.559 & 0.467 & \textbf{0.508}& 0.669 & 0.352 & 0.461         \\ \hline
  \textit{/organization/company}    & 7.8\% & 0.932 & 0.166 & \textbf{0.282}& 1.0   & 0.127 & 0.225         \\ \hline
  \textit{/location}                & 7.5\% & 0.687 & 0.796 & \textbf{0.737}& 0.787 & 0.609 & 0.687         \\ \hline
  \textit{/organization/government} & 2.1\% & 0     & 0     & 0             & 0     & 0     & 0             \\ \hline
  \textit{/location/country}        & 2.0\% & 0.783 & 0.614 & \textbf{0.688}& 0.838 & 0.498 & 0.625         \\ \hline
  \textit{/other/legal}             & 1.8\% & 0     & 0     & 0             & 0     & 0     & 0             \\ \hline
  \textit{/location/city}           & 1.8\% & 0.919 & 0.610 & \textbf{0.733}& 0.816 & 0.637 & 0.715         \\ \hline
  \textit{/person/political\_figure}& 1.6\% & 0     & 0     & 0             & 0     & 0     & 0             \\ \hline
\end{tabular}%
}
\caption{Performance analysis of the proposed model and AFET on top 10 (in terms of type frequency) types present in OntoNotes dataset.}
\label{tab:class-wise_OntoNotes}
\end{table}

\section{Conclusion and Future Work}
In this paper, we propose a neural network based model for the task of fine-grained entity classification. The proposed model learns representations for entity mention, its context and incorporate label noise information in a variant of non-parametric hinge loss function. Experiments show that the proposed model outperforms existing state-of-the-art models on two publicly available datasets without explicitly tuning data dependent parameters. 

Our analysis indicates the following observations. First, OntoNotes dataset has a different distribution of entity mentions compared with other two datasets. Second, if data distribution is similar, then transfer learning is very helpful. Third, incorporating data-driven label-label correlation helps in the case of labels of mixed types. Fourth, there is an inherent limitation in assuming all labels to be clean if they belong to the same path of the hierarchy. Fifth, the proposed model fails to learn label types that are very noisy.

Future work could analyse the effect of label noise reduction techniques on the proposed model, revisiting the definition of clean and noisy labels and modeling label-label correlation in a principled way that is not dependent on dataset specific parameters.

\section*{Acknowledgments}
We thank the anonymous reviewers for their invaluable and insightful comments. Abhishek is supported by MHRD fellowship, Government of India. We acknowledge the use of computing resources made available from the Board of Research in Nuclear Science (BRNS),  Dept\@. of Atomic Energy (DAE),  Govt\@. of  India  sponsered  project (No.2013/13/8-BRNS/10026) by Dr\@. Aryabartta Sahu at Department of Computer Science and Engineering, IIT Guwahati.
\bibliography{eacl2017}
\bibliographystyle{eacl2017}

\end{document}